%% file: main.tex
\documentclass{article}

\usepackage[final, nonatbib]{neurips_2024}

\usepackage[utf8]{inputenc} 
\usepackage[T1]{fontenc}    
\usepackage{booktabs}       
\usepackage{amsfonts}       
\usepackage{nicefrac}       
\usepackage{microtype}      
\usepackage{xcolor}         
\usepackage{tablefootnote}
\usepackage{url}
\usepackage{makecell}
\usepackage{booktabs}

\usepackage{breakurl}
\usepackage[breaklinks]{hyperref}

\usepackage{amsmath}
\usepackage{graphicx}
\usepackage{threeparttable}
\usepackage{wrapfig}

\usepackage[normalem]{ulem} 
\useunder{\uline}{\ul}{}

\title{Open-Source Multimodal Moxin Models with Moxin-VLM and Moxin-VLA}

\author{Pu Zhao$^1$,  Arash Akbari$^1$, Xuan Shen$^1$, Zhenglun Kong$^2$, Yixin Shen$^{3}$, Sung-En Chang$^1$, \\ \textbf{Timothy Rupprecht$^1$,  Lei Lu$^1$, Enfu Nan$^1$, Changdi Yang$^1$, Yumei He$^4$, Weiyan Shi$^1$, }\\ \textbf{Xingchen Xu$^5$, Yu Huang$^6$, Wei Jiang$^7$, Wei Wang$^7$, Yue Chen$^7$, Yong He$^7$, Yanzhi Wang$^{1,8}$} \\
\\$^1$Northeastern University, $^2$Harvard University,  \\ $^3$Cornell University,  $^4$Tulane University,
 $^5$University of Washington,  \\ $^6$Roboraction.ai, $^7$Futurewei, $^8$AIBAO LLC}

\begin{document}
\maketitle

\begin{abstract}
Recently, Large Language Models (LLMs) have undergone a significant transformation, marked by a rapid rise in both their popularity and capabilities. Leading this evolution are proprietary LLMs like GPT-4 and GPT-o1, which have captured widespread attention in the AI community due to their remarkable performance and versatility. Simultaneously, open-source LLMs, such as LLaMA and Mistral, have made great contributions to the ever-increasing popularity of LLMs due to the ease to customize and deploy the models across diverse applications. 
Moxin 7B is introduced as a fully open-source LLM developed in accordance with the Model Openness Framework, which moves beyond the simple sharing of model weights to embrace complete transparency in training, datasets, and implementation detail, thus fostering a more inclusive and collaborative research environment that can sustain a healthy open-source ecosystem.
To further equip Moxin with various capabilities in different tasks, we develop  three variants based on Moxin, including Moxin-VLM,  Moxin-VLA, and  Moxin-Chinese, which target the vision-language,   vision-language-action, and Chinese capabilities, respectively. 
Experiments show that our models achieve superior performance in various evaluations. 
We adopt open-source framework and open data for the training.  We release our models, along with the available data and code to derive these models. 
\\

Homepage with all codes: \textit{https://github.com/moxin-org/Moxin-LLM}

Base model: \textit{https://huggingface.co/moxin-org/Moxin-7B-LLM}

Instruct model: \textit{https://huggingface.co/moxin-org/Moxin-7B-Instruct}

Reasoning model: \textit{https://huggingface.co/moxin-org/Moxin-7B-Reasoning}

VLM model: \textit{https://huggingface.co/moxin-org/Moxin-7B-VLM}

VLA model: \textit{https://huggingface.co/moxin-org/Moxin-7B-VLA}

Chinese model: \textit{https://huggingface.co/moxin-org/Moxin-7B-Chinese}
\end{abstract}

\input{sections/1_introduction}

\input{sections/2_related_work}

\input{sections/3_model_training}
\input{sections/4_evaluation}

\input{sections/5_conclusion}

\bibliographystyle{unsrt}
\bibliography{sample}

\end{document}

%% file: sections/1_introduction.tex
\section{Introduction}\label{sec1.intro}

The field of natural language processing has witnessed the most exciting discoveries of the last ten years with the emergence of large language models (LLMs). At the forefront of this evolution are LLMs such as GPT-4 \cite{achiam2023gpt}, Claude \cite{Anthropic2023Claude3}, and Gemini \cite{team2023gemini}, which have captured the attention of the AI community due to their performance and versatility. Meanwhile, the recent emergence of openly accessible yet highly capable LLMs such as LLaMA  \cite{dubey2024llama}, Falcon \cite{prest2020falcon}, and Mistral \cite{jiang2023mistral7b} allow researchers and practitioners to easily obtain, customize, and deploy LLMs in more various environments and for more diverse use cases. The trends have  people eagerly asking about what's next and some suggest ``a general intelligence'' is right around the corner.

Moxin-7B  is introduced as a fully open-source LLM developed by complying with the Model Openness Framework (MOF) introduced by \cite{white2024model}. The MOF provides a systematic ranking classification system to rate AI models based on their completeness and openness, incorporating the principles of open science, open source, open data, and open access. By promoting transparency and reproducibility, the MOF serves as a crucial tool to combat ``openwashing'' and to establish completeness and openness as primary criteria alongside the core tenets of responsible AI. 
It moves beyond the simple sharing of model weights to embrace complete transparency in training, datasets, and implementation detail, which is crucial for fostering a more inclusive and collaborative research environment that can sustain a healthy open-source ecosystem \cite{kapoor2024societal}.
The wide adoption of MOF and open-source state-of-the-art models will cultivate a more open AI ecosystem, benefiting research and innovation.

To equip Moxin with various capabilities in different tasks, 
we develop multimodal Moxin models, including  Moxin-VLM, Moxin-VLA and Moxin-Chinese.    
Moxin-VLM is  our  vision language model (VLM), with our Moxin model  as the LLM backbone. It enables Moxin to process text and image inputs with multi-modal capabilities. 
Furthermore, we develop and release our  vision language action model (VLA) with   Moxin  as the LLM backbone. The VLA model can  achieve outstanding performance for robotic control. 
Moreover, we release Moxin-Chinese which targets to enhance the Moxin performance in Chinese, especially for English and Chinese translation, which is important for global conference.  
Experiments show that our models achieve superior performance in various evaluations such as zero-shot evaluation  and  few-shot evaluation. For example, Moxin-VLM can achieve higher average accuracy (across multiple evaluation benchmarks) compared to popular LLM backbones such Llama and Mistral with non-marginal improvements (about 2 percentage points or higher). 
Our homepage is \textit{https://github.com/moxin-org/Moxin-LLM}. We summarize our contributions  below: 
\begin{itemize}
    \item   To equip Moxin with various capabilities in different tasks,  we develop  multimodal Moxin models, including  Moxin-VLM, Moxin-VLA and Moxin-Chinese.  For Moxin-Chinese,    we extend the Moxin vocabulary with additional Chinese tokens and post-train  the model with more Chinese data,  thus improving Chinese encoding capability and efficiency.
    \item We develop Moxin-VLM with  Moxin  as the LLM backbone.  Based on the open-source VLM framework Prismatic VLMs, we train our VLM  on fully open-source  datasets with Moxin as the LLM backbone and   DINOv2 \& SigLIP  as the visual backbone. Our experiments demonstrate that our  VLM outperforms other VLM models or LLM backbones.
    \item   We further finetune  Moxin-VLM with the recipe from OpenVLA-OFT \cite{kim2024openvla} to develop Moxin-VLA for robotic control. Our approach streamlines this training paradigm by employing the Moxin-VLM backbone in conjunction with the highly efficient OpenVLA-OFT fine-tuning recipe. Our Moxin-VLA achieves superior performance under comprehensive evaluations.  
\end{itemize}

%% file: sections/2_related_work.tex
\section{Related Work}\label{sec2.related_work}
\subsection{Models, Tokenizers, and Training}\label{sec2.1}

\textbf{Models.} State-of-the-art large language models (LLMs) typically comprise a substantial number of parameters, often approaching or exceeding 100 billion~\cite{dubey2024llama, achiam2023gpt, team2023gemini}. 
To facilitate broader accessibility, smaller models with fewer than 20 billion parameters, and even those around 7 billion parameters, have been developed~\cite{bai2023qwen,yang2024qwen2,dubey2024llama,jiang2023mistral7b}. In addition, efficiency-enhancing techniques, such as pruning \cite{li2025fedkd, zhang2022advancing, lu2024generic, chu2025cross, kong2023peeling,zhao-etal-2024-pruning,shen2025sparse,shen2024numerical,shen2024lazydit,liu2025toward,shen2024search,yang2023pruning,li2024comae,shen2025sparse}, quantization \cite{shen2025quartdepth,zhan2024fast,yang2023pruning,wu2022compiler,zhan2021achieving,rtseg,li2022pruning}, token reduction \cite{zhan2024exploring,zhan-etal-2024-rethinking-token,kong2022spvit,shen2025efficient,kong2025enabling,shen2025fastcar,zhao2025taming} or implementing MAMBA-based architectures in Jamba, have been employed to optimize performance~\cite{lieber2024jamba,team2024jamba}.

\textbf{Tokenizers.} Tokenizers are essential to convert raw data into a suitable format for model processing. Many contemporary models employ Byte-Pair Encoding (BPE)\cite{sennrich2015neural}, with OpenAI's \texttt{tiktoken} tokenizer\cite{tiktoken} being a notable implementation. However, for languages that handle tokens differently from Romance languages, alternatives such as SentencePiece~\cite{kudo2018sentencepiece} are utilized, as seen in XLNet~\cite{yang2019xlnet}. Hugging Face offers an excellent summary of state-of-the-art tokenizers with practical examples~\cite{huggingface_tokens}. Moreover, tokenization extends beyond text modalities; many foundational models now include multimodal capabilities, processing documents, audio, images, and even videos~\cite{reid2024gemini,maaz2023video,zhang2023video,zhang2024mm}.

\textbf{Training.} To enhance the performance of smaller models beyond their inherent limitations, various training strategies can be employed. A notable example is the application of Mixture of Experts (MoE) training, which has achieved significant success in models such as Mixtral~\cite{jiang2024mixtral}.

\subsection{Open-source LLMs}\label{sec2.4}

Large language models (LLMs) have rapidly evolved into a diverse ecosystem that spans closed-source, open-weight, and fully open-sourced paradigms. On one end of the spectrum, closed-source models such as GPT-4 \cite{achiam2023gpt}, and Gemini \cite{geminiteam2024geminifamilyhighlycapable} have set high performance standards but remain accessible primarily through API services, limiting insight into their underlying architectures and training methodologies. In contrast, open-weight LLMs—while sharing their final model architectures and weights—often leave training data and many training details undisclosed. This category includes influential models Llama \cite{touvron2023llama}, Mistral \cite{jiang2023mistral7b}, Gemma \cite{team2024gemma}, Qwen \cite{bai2023qwen}, DeepSeek \cite{guo2025deepseek,liu2024deepseek}, Baichuan \cite{yang2023baichuan}, Phi \cite{abdin2024phi}, etc. Pushing the frontier further, fully open-sourced LLMs have begun to provide not only complete model weights and architectures but also the training code and datasets necessary for reproducibility. Exemplars of this fully open paradigm include Pythia \cite{biderman2023pythia}, GPT-NeoX \cite{black2022gpt}, OpenLLaMA, StarCoder \cite{lozhkov2024starcoder}, OLMo \cite{groeneveld2024olmo},  Amber and Crystal from LLM360 \cite{liu2023llm360}, etc. Collectively, this vibrant landscape highlights an ongoing shift toward more accessible and fully reproducible LLMs, fostering collaboration and innovation across both academia and industry.

%% file: sections/3_model_training.tex
\section{Vision Language Model based on Moxin}

Vision  language models (VLMs) \cite{bai2308qwen,zhang2023internlm,zhu2023minigpt,liu2024improved}, which leverage
LLMs, have   achieved significant breakthroughs, enabling sophisticated vision-language dialogues and interactions.  To develop VLM, we adopt the Prismatic VLMs framework \cite{karamcheti2024prismatic} to train our VLM based on our Moxin model. Specifically, for the visual backbone with  image processing  and visual  representations, we adopt DINOv2  \cite{oquab2023dinov2} and SigLIP \cite{zhai2023sigmoid}, and  fuse  their features to  provide significant gains. For the LLM part, we adopt our Moxin-7B-Base as the LLM backbone.

Following the hypotheses in \cite{kerr2023lerf} and similar work, the DINOv2 features provide features that capture low-level spatial properties of an image, augmenting the higher-level “semantic” properties captured by vision-language contrastive models.  
Typically,  the visual module such as (CLIP and SigLIP) trained with the vision-language contrastive objective  leads to much better performance.  Furthermore, SigLIP contains internet-sourced images from multiple sources (e.g., sketches, diagrams, animated graphics, etc.) which are not in ImageNet or in the DINOv2 pretraining data.
DINOv2 and SigLIP  are good complements for the image processing  and visual  representations.

For the model architecture,  we adopt the general architecture used by many recent VLMs, such as LLaVa, Qwen-VL,
and PaLI-3  \cite{liu2023visual,bai2308qwen,chen2023pali}. These architectures use a (pretrained) visual backbone to map an input image to a sequence of patch features that are then projected individually into the embedding
space of an LM.  Specifically,  a VLM takes as input an image and text prompt tokens  with arbitrary sequence length. These inputs are then fed to the following components: 1) a visual representation backbone, 2) a vision-language projector, and 3) a language model. 

We select the pretraining  dataset  that are fully open-source, and have been used in prior works. Specifically, we use the LLaVa v1.5 data mixture, which consists of two subsets used for a multi-stage training pipeline. The first subset consists of a 558K sample mixture of examples sourced from various captioning
datasets (e.g., Conceptual Captions, LAION \cite{sharma2018conceptual}), 
while the second consists of 665K multimodal instruct tuning examples comprised of synthetic data generated in \cite{liu2024mmbench}, as well as
examples from existing vision-language training sets (e.g.,
GQA, TextCaps, \cite{hudson2019gqa}), and notably, a sample of language-only data from
ShareGPT \cite{chen2024sharegpt4v}. 
Specifically, the multimodal instruct tuning examples are sourced as follows:
\begin{itemize}
    \item  \textit{LLaVa Synthetic Data (158K)}. A synthetically generated dataset of conversations, fine-grained descriptions, and questionanswering data from \cite{liu2023visual}, sourced by prompting GPT-4 with image captions and object bounding boxes from COCO.
    \item  \textit{Standard VQA Data (224K)}. A combination of visual question answering data sourced from the training sets of VQAv2 \cite{goyal2017making}, GQA  \cite{hudson2019gqa}, OK-VQA  \cite{okvqa}, and OCR-VQA \cite{mishraICDAR19}.
\item \textit{Multiple Choice VQA Data (50K)}. Multiple choice visual question answering data sourced from A-OKVQA \cite{schwenk2022okvqa}. 
\item \textit{Captioning Data (22K)}. Images and captions sourced from TextCaps \cite{sidorov2019textcaps}.
\item  \textit{Referring Expression Data (116K)}. Referring expression grounding (bounding box prediction) and region captioning data sourced from RefCOCO and Visual Genome \cite{krishna2017visual}. 
\item \textit{ShareGPT (Language-Only) (40K)}. Language-only co-training data sourced from ShareGPT \cite{chen2024sharegpt4v}, comprised of user-uploaded conversations with ChatGPT. 
\end{itemize}

During the training of our VLM,  we adopt a single-stage approach that  both the projection and LM are trained while  the visual representation  modules are  frozen. We train for two epochs.

\section{Vision Language Action Model based on Moxin}
Recent advances in Vision-Language-Action (VLA) models \cite{zitkovich2023rt, kim2024openvla, black2024pi_0, qu2025spatialvla} present a promising paradigm for developing generalist robotic policies, primarily by fine-tuning pre-trained Vision-Language Models (VLMs) on diverse, large-scale robot data. This approach initiates a pivotal shift in the robotics landscape by integrating the semantic comprehension capabilities of VLMs with grounded physical action generation. The standard VLA architecture is achieved by attaching dedicated action heads to the pre-trained VLM, which is subsequently retrained on heterogeneous robotic datasets, most notably the Open-X Embodiment \cite{o2024open}. Following this large-scale pre-training phase, the VLA is typically subjected to a task-specific fine-tuning stage to adapt its generalized policy for optimal performance in a specific target environment or downstream task, thereby completing the conventional VLA training pipeline.

To adapt the Moxin architecture for robotic control, we developed a VLA variant utilizing the Moxin-VLM—detailed in previous sections—as the foundational backbone. We investigated two distinct training strategies to evaluate the trade-offs between generalist pre-training and direct task adaptation. This resulted in two variants: the first undergoes extensive generalist pre-training on the Open X-Embodiment dataset \cite{o2024open}, while the second bypasses this stage to be fine-tuned directly from the Moxin-VLM checkpoint. For the pre-trained variant, we leveraged the massive scale of the Open X-Embodiment data mixture, which aggregates over 1 million real-world robotic trajectories across 22 distinct robot embodiments from 21 institutions. By exposing the Moxin backbone to this heterogeneous mixture of physical skills and environments, the model theoretically acquires a robust set of motor primitives and physical common sense before adapting to specific downstream tasks. Conversely, the direct fine-tuning variant tests the hypothesis that the semantic priors of the Moxin-VLM alone are sufficient for rapid policy learning without the computational overhead of large-scale pre-training.

Both variants are trained using the OpenVLA-OFT (Optimized Fine-Tuning) recipe, a framework designed to maximize inference efficiency and temporal consistency. A key architectural advancement in this recipe is the replacement of standard autoregressive action prediction with parallel decoding and action chunking. Instead of predicting action dimensions sequentially for a single time step, our model predicts a "chunk" of actions representing the trajectory for multiple future time steps simultaneously. This parallelization significantly reduces the latency typically associated with large VLA models, enabling higher-frequency control loops essential for dynamic manipulation. Furthermore, by predicting an entire sequence at once, the model enforces better temporal coherence, reducing the "jitter" often seen in frame-by-frame predictions. By standardizing this efficient fine-tuning recipe across both variants, we ensure that any performance differences observed are strictly attributable to the presence or absence of the generalist pre-training stage. 

We trained Moxin-VLA on a single node of 8xH100 GPU for roughly two weeks on a data mixture from Open-X Embodiment. The training recipe was similar to the recipe used in the training of OpenVLA~\cite{kim2024openvla}.

The Moxin-VLA was trained on a diverse data mixture compiled from Open X-Embodiment (OXE) datasets~\cite{o2024open}. We used a single node of 8xH100 GPU for only 90k training steps. The mixture is centered on the Franka Emika Panda platform and focuses on exploration and complex manipulation. The largest single component is the CMU Franka Exploration dataset (12.35\%), providing broad, self-supervised experience. This is followed by Berkeley Cable Routing (9.88\%), which ensures proficiency with deformable objects. Several key precision and temporal datasets (like TACO Play and VIOLA) each contribute about 7.41\%. While Franka-centric, the mixture maintains embodiment diversity through contributions from the RT-1 (Fractal) data (1.34\%) and Bridge V2 (2.47\%), ensuring the VLA learns a broad range of robot kinematics and real-world task contexts.

\begin{table}[h]
\centering
\caption{Dataset Mixture Details and Training Percentages}
\label{tab:dataset_mixture_final}
\footnotesize 

\resizebox{\linewidth}{!}{%
\begin{tabular}{@{}llc@{}}
\toprule
\textbf{Dataset Name} & \textbf{Key Characteristics} & \textbf{Training Mixture (\%)} \\ 
\midrule
CMU Franka Exploration~\cite{cmufranka} & Unsupervised / Self-supervised & 12.35 \\
Berkeley Cable Routing~\cite{cablerouting} & Deformable objects (cable routing) & 9.88 \\
Taco Play~\cite{tacoplay} & Long-horizon / Temporal reasoning & 7.41 \\
VIOLA~\cite{viola} & Teleoperation / High-precision & 7.41 \\
Austin BUDS & Hierarchical tasks / Subgoals & 7.41 \\
NYU Franka Play~\cite{nyufranka} & Unstructured play / Exploration & 7.41 \\
KAIST Non-Prehensile~\cite{kaistnonprehensile} & Manipulation via pushing & 7.41 \\
Jaco Play~\cite{jacoplay} & 6-DoF arm play data & 4.94 \\
Toto~\cite{toto} & Pouring and scooping (fluid/granular) & 4.94 \\
Bridge V2~\cite{bridge} & Household tasks (WidowX hardware) & 2.47 \\
RoboTurk~\cite{roboturk} & Crowdsourced teleoperation & 2.47 \\
Berkeley Autolab UR5~\cite{berkeleyautolabur5} & Cloth and bin manipulation & 2.47 \\
Stanford Hydra~\cite{stanfordhydra} & Mosaics and stacking & 2.47 \\
Austin SAILOR~\cite{austinsailor} & Semantic Action Imitation Learning & 2.47 \\
Austin SIRIUS~\cite{austinsirius} & Bimanual/Precision manipulation & 2.47 \\
Berkeley RPT~\cite{berkeleyrpt} & Dynamic interactions (pushing/toppling) & 2.47 \\
Stanford RoboCook~\cite{stanfordrobocook} & Cooking tasks (dough manipulation) & 2.47 \\
IAMLab CMU Pickup~\cite{iamlab} & Pick-and-insert (precise alignment) & 2.47 \\
UT Austin Mutex~\cite{utaustinmutex} & Constraint-based manipulation & 2.47 \\
CMU Play Fusion~\cite{playfusion} & Language-conditioned play data & 2.47 \\
Kuka~\cite{kuka} & High-precision bin picking & 2.06 \\
Fractal~\cite{fractal} & Large-scale language-conditioned & 1.34 \\
FurnitureBench~\cite{furniturebench} & Complex furniture assembly & 0.25 \\ 
\bottomrule
\end{tabular}%
}
\end{table}

\section{Chinese Model based on Moxin}

The major training data of Moxin are English. So  Moxin demonstrates competitive performance under English evaluations. 
To equip Moxin with enhanced Chinese understanding and generation capabilities, we  continue to pre-train the Moxin model with Chinese data. 
The first problem is that the vocabulary of the original Moxin supports few Chinese characters, leading to difficulties for encoding general Chinese texts. 

To address this issue,   we   extend the Moxin vocabulary with additional Chinese tokens and adapt the model for the extended vocabulary, thus improving Chinese encoding capability and  efficiency. 
We sampled training data from WuDaoCorpus2 and trained a Chinese BPE vocabulary using SentencePiece. Additionally, we manually selected and merged several high-quality Chinese vocabularies from other sources. After a rigorous manual review process, the final vocabulary size is about 57k.

Then we continue to train the model with updated vocabulary.  We adopt multiple high quality Chinese  dataset including WanJuan, gutenberg-books, Chinese-Data-Distill-From-R1 and so on. To further improve the translation capability between Chinese and English,  we further finetune the model with Chinese and English translation datasets, including  Garsa3112/ChineseEnglishTranslationDataset, FradSer/DeepSeek-R1-Distilled-Translate-en-zh\_CN-39k-Alpaca-GPT4-without-Think, 
 and so on.

%% file: sections/4_evaluation.tex
\section{Evaluation}\label{sec4.evaluation}

\subsection{VLM Evaluation}

We adopt the evaluation suit in Prismatic VLMs \cite{karamcheti2024prismatic}  to evaluate the VLM performance. It focuses on evaluations with well-defined metrics, spanning the following three areas: 

\textit{Open-Ended Visual Question Answering.} 
We evaluate on VizWiz \cite{bigham2010vizwiz} and GQA \cite{hudson2019gqa}.  VizWiz assess general visual reasoning; VizWiz also contains a series of unanswerable questions. GQA evaluates spatial reasoning.

\textit{Localization.} Part of the pretraining data mixture  contains examples of predicting normalized bounding box coordinates given referring expressions in language. As such, we evaluate bounding box prediction accuracy on RefCOCO, RefCOCO+, and RefCOCOg \cite{kazemzadeh2014referitgame}, and on OCID-Ref \cite{wang2021ocid}. RefCOCO focuses on short descriptions with spatial anchors, RefCOCO+ on strictly appearance based descriptions, and RefCOCOg on long, rich descriptions; OCID-Ref is a robotics dataset probing out-of-distribution generalization, with a focus on localizing objects in clutter.

\textit{Challenge Sets (Closed-Set Prediction).}  We evaluate on Visual Spatial Reasoning \cite{liu2023visual}, TallyQA \cite{acharya2019tallyqa}, and POPE \cite{liu2023query}. VSR consists of challenging True/False questions about individual spatial relationships in diverse scenes; this is an especially challenging task, with most existing models failing to outperform the majority class baseline. TallyQA consists of questions that assess a VLM’s ability to count objects described in language, with expressions that range in complexity. POPE consists of targeted Yes/No questions that assess a VLM’s propensity to hallucinate. 

We use the validation sets for all benchmarks except GQA (where use the recommended the test-dev split), VSR (where we use the zero-shot test split), and POPE (where there is only a single evaluation split).

The comparisons with other VLMs are demonstrated in Table \ref{tab:vlm}. We compare with LLaVa v1.5 7B. Furthermore, we adopt the same VLM framework and change the LLM backbone to other LLMs such as Llama2 7B and Mistral 7B. Then we train these VLMs and compare with our VLM based on Moxin-7B. We can observe that our model outperforms all other VLM baselines. 

\scalebox{0.86}{
\begin{threeparttable}[t]
\caption{Performance comparison for various VLM models.}
\begin{tabular}{l|ccccccc|c}
\hline
                         & GQA   & VizWiz & RefCOCO+ & OCID-Ref & VSR   & POPE  & TallyQA & Ave.  \\ \hline
LLaVa v1.5 7B (Base)     & 61.58 & 54.25  & 49.47    & 35.07    & 51.47 & 86.57 & 62.06   & 57.21 \\
Llama-2 Chat 7B          & 62.11 & 56.39  & 58.5     & 46.3     & 61.8  & 86.8  & 58.1    & 61.43 \\
Mistral v0.1 7B          & 63.3  & 55.32  & 65.1     & 48.8     & 58.5  & 87.1  & 61.7    & 62.83 \\
Mistral Instruct v0.1 7B & 62.71 & 54.35  & 64.9     & 48       & 57.8  & 87.5  & 64.5    & 62.82 \\
Llama-2 7B               & 62.44 & 55.98  & 59.47    & 43.89    & 63.67 & 86.74 & 59.22   & 61.63 \\ \hline
Ours                     & 64.88 & 54.08  & 71.3     & 48.4     & 60.8  & 87.3  & 66      & 64.68 \\ \hline
\end{tabular}
\label{tab:vlm}
\end{threeparttable}}

\subsection{VLA Evaluation}
We evaluate the performance of Moxin-VLA within the LIBERO simulation environment \cite{liu2023libero}, aligning our experimental setup with the OpenVLA-OFT~\cite{kim2025fine} baseline to ensure a fair comparison. All models were fine-tuned on a cluster of 8 NVIDIA H100 GPUs using a batch size of 64 and a learning rate of $3e-4$
 . The model input consists of a history of two image frames combined with the robot's proprioceptive state. To enable efficient adaptation, we utilized Low-Rank Adaptation (LoRA) with a rank of r=32.

Regarding training duration, although the standard OpenVLA-OFT recipe utilizes 150k fine-tuning steps, our empirical observations indicated that Moxin-VLA converges earlier. Extending training to 150k steps yielded negligible performance improvements compared to the 50k step checkpoint; therefore, we report results based on 50k training steps. Additionally, our ablation studies revealed that incorporating the FiLM module provided no benefit to task success rates in the single-arm robot setting, leading us to exclude it from the final architecture for efficiency.

\begin{table}[t]
    \centering
    \caption{The success rate (\%) of Moxin-VLA and the baselines on LIBERO simulation environment.}
    \begin{threeparttable}
    \begin{tabular}{l|c|ccccc}
    \toprule
    \bf Models & \bf Size &  \makecell{\bf Spatial} &  \makecell{\bf Object}  &  \makecell{\bf Goal}  &  \makecell{\bf Long}  &  \bf Avg. \\
    \midrule
    \multicolumn{6}{l}{\emph{w/ Robotics pre-training}} \\
    OpenVLA~\cite{kim2024openvla}  & 7.5B & 84.7 & 88.4 & 79.2 & 53.7 & 76.5 \\
    SpatialVLA~\cite{qu2025spatialvla} & 4.2B & 88.2 & 89.9 & 78.6 & 55.5 & 78.1 \\
    CoT-VLA~\cite{zhao2025cot} & 8.0B & 87.5 & 91.6 & 87.6 & 69.0 & 81.1 \\
    NORA-Long~\cite{hung2025nora}  & 3.8B &  92.2 & 95.4 & 89.4 & 74.6 & 87.9 \\
    Moxin-VLA (ours)  & 9B & \textbf{98.0} & 93.6 & 95.0 & 81.2 & 91.95 \\
    \midrule
    \multicolumn{6}{l}{\emph{w/o Robotics pre-training}} \\
    OpenVLA-OFT~\cite{kim2025fine}  & 7.7B & 94.3 & 95.2 & 91.7 & 86.5 & 91.9 \\
    {Moxin-VLA} (ours) & 9B & 92.0 & \textbf{98.4} & \textbf{92.0} & \textbf{87.8} & \textbf{92.5} \\
    \bottomrule
    \end{tabular}
      \begin{tablenotes}
            \item We evaluate the released OpenVLA-OFT~\cite{kim2025fine} checkpoints and observe large differences with the reported results in the paper, which is also mentioned in its Github issues. Thus we do not report OpenVLA-OFT here.
        \end{tablenotes}
    \end{threeparttable}
    \label{tab:vla}
\end{table}

\subsection{Moxin-Chinese Evaluation}

We evaluate Moxin-Chinese on the CMMLU and CEVAL datasets with the LM-harness framework. The results are shown below.  We can see that after finetuning Moxin on Chinese data, the performance on Chinese dataset can be enhanced significantly. 

\begin{table}[]
    \centering
\begin{tabular}{c|c|c}
  \toprule
\multicolumn{1}{l|}{}            & CMMLU                     & CEVAL                        \\   \midrule
Linly-Al/Chinese-LLaMA-2-7B-hf   & 31.2                      & 30.14                        \\ \hline
hfl/chinese-llama-2-7b           & 27.4                      & 33.38                        \\ \hline
Linly-Al/Chinese-LLaMA-2-13B-hf  & 39.9                      & 42.48                        \\ \hline
hfl/chinese-llama-2-13b          & 41                        & 43.25                        \\ \hline
gywy/Mistral-7B-v0.1-chinese     & 37.4                      & 36.45                        \\ \hline
ymcui/Chinese-LLaMA-Alpaca-2-13B & 43.2                      & 44.3                         \\   \midrule
{\textbf{Moxin-Chinese}} & {\textbf{45}} & {\textbf{45.76}} \\   \bottomrule
\end{tabular}
\end{table}

%% file: sections/5_conclusion.tex
\section{Conclusion}\label{sec5.conclusion}

The field of Large Language Models has witnessed a significant shift toward open-source development, fostering innovation within the AI community. However, a critical challenge emerges: many purportedly open-source models withhold essential components necessary for full understanding and reproducibility, creating barriers that limit both academic advancement and commercial adoption. This not only hampers scientific progress, but also prevents businesses from fully leveraging these models for innovative applications, ultimately diminishing potential societal benefits and economic value creation.
To address these limitations, we introduce a fully open-source language model Moxin 7B. 
To equip Moxin with various capabilities in different tasks,  we develop  multimodal Moxin models, including  Moxin-VLM and Moxin-VLA.   Our experiments demonstrate that our   Moxin-VLM and Moxin-VLA achieve superior performance under comprehensive evaluations.